\documentclass[letterpaper]{article} 
\usepackage{aaai24}  
\usepackage{times}  
\usepackage{helvet}  
\usepackage{courier}  
\usepackage[hyphens]{url}  
\usepackage{graphicx} 
\usepackage{natbib}  
\usepackage{caption} 
\usepackage{array,booktabs,siunitx}
\usepackage{tabularx}

\usepackage{algorithm}
\usepackage{algorithmic}

\usepackage{graphicx}
\usepackage{multirow}
\usepackage{multicol}
\usepackage{amsmath,amsfonts,bm}
\usepackage{enumitem}
\usepackage{amssymb}
\usepackage{colortbl}

\usepackage{xcolor}
\definecolor{myblue}{RGB}{218,232,252}

\usepackage{newfloat}
\usepackage{listings}
\DeclareCaptionStyle{ruled}{labelfont=normalfont,labelsep=colon,strut=off} 
\floatstyle{ruled}
\newfloat{listing}{tb}{lst}{}
\floatname{listing}{Listing}
\title{LLM-BIP: Structured Pruning for Large Language Models with Block-Wise Forward Importance Propagation}
\author{
    Haihang Wu
}
\affiliations{
    The University of Melbourne\\

}

\usepackage{bibentry}

\begin{document}

\maketitle

\begin{abstract}
Large language models (LLMs) have demonstrated remarkable performance across various language tasks, but their widespread deployment is impeded by their large size and high computational costs. Structural pruning is a
prevailing
technique used to introduce sparsity into pre-trained models and facilitate direct hardware acceleration during inference by removing redundant
connections (structurally-grouped parameters), such as channels and attention heads. Existing structural pruning approaches
often employ either global or layer-wise pruning criteria; however, they are hindered by ineffectiveness stemming from inaccurate evaluation of connection importance.
Global pruning methods typically assess component importance using near-zero and unreliable gradients, while layer-wise pruning approaches encounter significant pruning error accumulation issues. To this end, we propose a more accurate pruning metric based on the block-wise importance score propagation, termed LLM-BIP. Specifically, LLM-BIP precisely evaluates connection importance by gauging its influence on the respective transformer block output, which can be efficiently approximated in a single forward pass through an upper bound derived from the assumption of Lipschitz continuity.
We evaluate the proposed method using LLaMA-7B, Vicuna-7B, and LLaMA-13B across common zero-shot tasks. 
The results demonstrate that our approach achieves an average of 3.26\% increase in accuracy for common reasoning tasks compared to previous best baselines.
It also reduces perplexity by 14.09 and 68.76 on average for the WikiText2 dataset and PTB dataset, respectively.
\end{abstract}

\section{Introduction}

Large language models (LLMs) have demonstrated exceptional performance across various language tasks such as question answering ~\cite{Sakaguchi2019WinoGrande:Scale,Zellers2019HellaSwag:Sentence,Clark2018ThinkChallenge} and language generation~\cite{Wei-LinChiang2023Vicuna:Quality,Touvron2023LLaMA:Models}. However, this notable performance is accompanied by significant computational requirements and model scale, leading to high memory costs and latency. To mitigate these challenges, techniques including knowledge distillation~\cite{Sun2019PatientCompression,Sun2020ContrastiveCompression}, quantization~\cite{Frantar2023GPTQ:Transformers,Dettmers2023SpQR:Compression} and network pruning~\cite{Zhuang2018Discrimination-awareNetworks, Xia2023ShearedPruning} have been employed to compress 
LLMs.
Among these methods, structural network pruning has emerged as a particularly effective method in compressing models and reducing latency by removing redundant structurally grouped parameters, such as attention heads and channels, from a pre-trained network, enabling direct inference speedup without necessitating specialized hardware support~\citep{Fang2023DepGraph:Pruning,Yang2023GlobalSaliency}. Existing structural pruning approaches typically prune networks either globally~\citep{Ma2023LLM-Pruner:Models,Zhang2023LoRAPrune:Fine-Tuning, Yu2018NISP:Propagation,Frantar2023SparseGPT:One-Shot} or layer-wisely~\cite{Sun2024AModels,Frantar2023SparseGPT:One-Shot}. 

\begin{figure*}[t]
\begin{center}
   \includegraphics[width=0.85\linewidth]{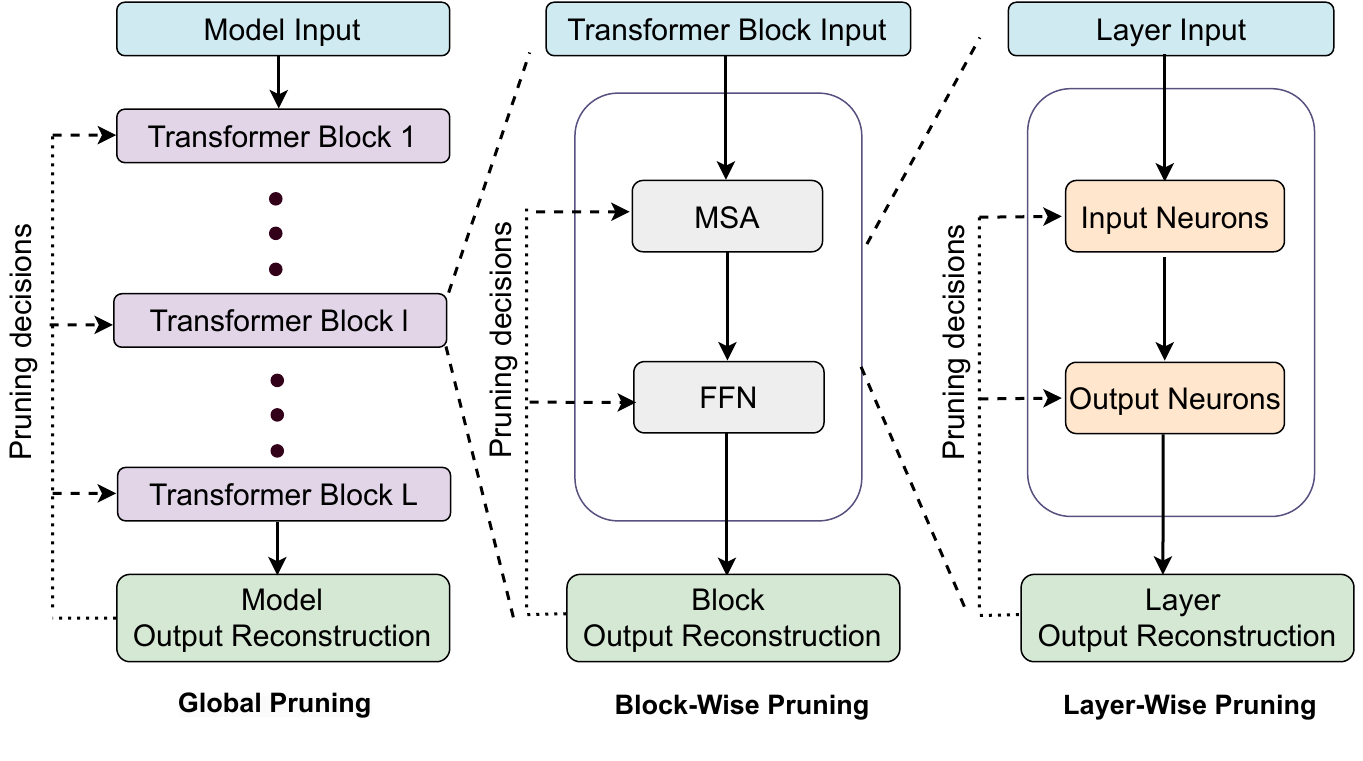}
\end{center}
\vspace{-1em}
   \caption{The comparison among global pruning, layer-wise pruning, and our block-wise pruning. Global pruning methods~\cite{Ma2023LLM-Pruner:Models} target on the minimization of the pruning effects on the final model output, typically relying on the memory-intensive and unreliable gradients. Layer-wise pruning techniques~\cite{Sun2024AModels} focus on pruning error minimization on the output of the current layer. 
   Despite its pruning efficiency, it suffers from the rapid pruning error accumulation issue.
   In contrast, our block-wise pruning strategy aims to minimize the pruning impact on the output of the transformer block, avoiding unreliable gradients and mitigating the error accumulation issue.}

\label{fig: overview}
\end{figure*}

Global structural pruning approaches (Figure ~\ref{fig: overview}) prune grouped parameters with minor impacts on the network's final output~\cite{Ma2023LLM-Pruner:Models,Zhang2023LoRAPrune:Fine-Tuning,Tang2022PatchTransformers}. For some global pruning approaches, they evaluate the importance of parameters based on their gradients, and remove structures with small gradients~\cite{Ma2023LLM-Pruner:Models,Zhang2023LoRAPrune:Fine-Tuning}. However, 
the gradient criterion 
can be unreliable due to small scales in well-trained models~\cite{Cun1989OptimalDamage,Frantar2023SparseGPT:One-Shot}, and also entails substantial computation and memory costs of these gradients particularly in LLMs. With the Lipschitz continuity assumptions on network components (e.g., activation function, linear projection layers), other global pruning approaches~\cite{Tang2022PatchTransformers,Yu2018NISP:Propagation} assess neuron importance by propagating neuron importance scores from the final layer throughout the network, and prune structures with small importance scores. By pruning structures deemed unimportant to the network output, the network's function may be preserved effectively. However, Lipschitz continuity assumption is proved to be invalid for common dot-product self-attention module in LLM transformer blocks~\cite{Kim2021TheSelf-Attention}.

Conversely, layer-wise pruning approaches remove grouped parameters with minimal impacts on the current layer's output~\cite{Sun2024AModels,Frantar2023SparseGPT:One-Shot}, typically requires no backpropagation. SparseGPT~\cite{Frantar2023SparseGPT:One-Shot} utilizes optimal brain surgeon to identify and prune unimportant weights. The remaining weights are then updated to reconstruct the output of the current layer. Wanda~\cite{Sun2024AModels} eliminates unimportant weights based on their magnitude and input feature norm of the current layer.  By pruning parameters
layer-wisely, the pruning efficiency is promising as only one forward pass is needed~\cite{Sun2024AModels}. 
However, layer-wise pruning can suffer from rapid error accumulation issues across layers~\cite{Xu2024BESA:Allocation}.

To address these issues, we propose an effective and efficient block-wise importance score propagation strategy for pretrained LLMs, termed LLM-BIP. LLM-BIP aims to preserve the  response of a transformer block as much as possible after pruning channels and attention heads in this block. With binary vectors representing pruning decisions, we formulate this objective as a mathematical optimization problem: searching binary vectors to minimize the $l_1$ distance between the output produced by the original network and a pruned network. As the searching cost is high due to the large searching space, we introduce a relaxation of this objective and derive an upper bound for this $l_1$ distance. Remarkably, minimizing this upper bound yields a straightforward metric for evaluating channel importance. Leveraging this metric, we prune network channels within the Feed-Forward Network module, and attention heads of the self-attention module based on their importance scores aggregated from the importance scores of  corresponding output channels. Notably, this also allows for the efficient pruning of the network within \textit{a single forward process}. Our approach enhances pruning accuracy by circumventing the reliance on inaccurate gradients~\cite{Cun1989OptimalDamage,Frantar2023SparseGPT:One-Shot} and the invalid Lipschitz continuity assumption on self-attention modules~\cite{Kim2021TheSelf-Attention}, which are employed by global pruning methods. Additionally, our method evaluates channel importance on a broader scale, mitigating the rapid error accumulation issue associated with existing layer-wise pruning approaches~\cite{Sun2024AModels,Frantar2023SparseGPT:One-Shot}.

We assess the efficacy of our proposed method on LLaMA-7B, Vicuna-7B, and LLaMA-13B using standard zero-shot datasets. Our experiments demonstrate that compared to Wanda~\cite{Sun2024AModels} and LLM-Pruner~\cite{Ma2023LLM-Pruner:Models}, our method reduces perplexity by 14.09 and 68.76 on average for WikiText2 dataset and PTB dataset respectively.  It also yields an average 3.26\% increase in accuracy for common reasoning tasks. 

In summary, our contributions are:

\begin{itemize}
    \item  
    We propose a simple and effective block-wise structured pruning criterion, aimed at diminishing error accumulation and eradicating reliance on gradients. By deriving an upper bound to approximate the impact of the attention heads and FFN channels on the block output, we enable a single forward pass for efficient importance estimation.
    
    \item  Our approach outperforms existing structured pruning methods across various models, achieving an average perplexity reduction of 3.67 and 88.63 at 20\% and 50\% sparsities on WikiText2 and PTB datasets. We also observe an average accuracy improvement of 4.49\% and 2.18\% at 20\% and 50\% sparsities on seven common-sense reasoning datasets.
\end{itemize}

\section{Related Work}

Existing pruning methods
can be categorized into unstructural pruning~\cite{Dong2017LearningSurgeon,Lee2019SNIP:Sensitivity,Singh2020WoodFisher:Compression,Evci2020RiggingWinners,Jin2022PruningsRegularization,Zhang2022PLATON:Importance}, semi-structural pruning~\cite{Zhou2021LEARNINGSCRATCH,Kurtic2022TheModels}, and structural pruning~\cite{Zhuang2018Discrimination-awareNetworks, Xia2023ShearedPruning,Nova2023Gradient-FreeData,Liu2023DejaTime,Fang2023DepGraph:Pruning,Yang2023GlobalSaliency}. Unstructured pruning approaches identify less important weights and zeroing out these weights without architecture change. Semi-structured pruning only stores $N$ nonzero weights in  each contiguous block of $M$ values for $N:M$ fine-grained structured sparsity pattern~\cite{Zhang2022LearningSparsity,Pool2021ChannelSparsity}. This accelerates the in- ference speed by only accessing and operating on these nonzero weights. However, both unstructured and semi-structured pruning typically require specialized AI accelerators or software to translate the compression into tangible inference latency savings. By contrast, structural pruning approaches remove structurally-grouped parameters such as channels~\cite{Hou2022CHEX:Compression,Shi2023UPop:Transformers}, tokens~\cite{Chen2023DiffRateTransformers,Kong2022SPViT:Pruning}, attention heads~\cite{Chen2021ChasingExploration,Kurtic2023ZipLM:Models} to achieve direct speedup, which is friendly to general hardware platforms. For this reason, our study focuses on structural pruning.

The structural pruning of pretrained models encompasses layerwise pruning~\cite{Dong2017LearningSurgeon,Jiang2018EfficientError,Frantar2022OptimalPruning, Sun2024AModels} and global pruning~\cite{Molchanov2019ImportancePruning,Yu2022WidthTransformers,Tang2022PatchTransformers}. Layerwise pruning methods assess the importance of parameters layerwisely based on their effect on the output  of the same layer. These layerwise pruning approaches often need only one forward process to prune the network. To improve efficiency further, Wanda introduces a metric based on weight magnitude and input feature norm to alleviate the SparseGPT's high computational burden associated with the Hessian matrix. Although efficient,  these approaches~\cite{Sun2024AModels,Frantar2023SparseGPT:One-Shot} suffer from rapid error accumulation issue~\cite{Xu2024BESA:Allocation}. Conversely, global pruning approaches evaluate the importance of parameters by analyzing their influence on the network output. However, their evaluation is based on either small and memory-intensive gradients ~\cite{Hassibi1993OptimalPruning, Yu2022TheNetworks} or invalid Lipschitz continuity assumption on self-attention modules~\cite{Kim2021TheSelf-Attention,Tang2022PatchTransformers}. Additionally, global pruning approaches are often less efficient than layer-wise approaches, usually requiring one forward process and one backward process. Different from the previous methods, our proposed structured pruning criterion is more efficient and accurate as only a single forward pass is needed without replying on gradients.

Block-wise compression methods have been developed for both quantization~\cite{Bai2022TowardsModels} and unstructured pruning~\cite{Xu2024BESA:Allocation}. These approaches minimize reconstruction error by either directly training quantized blocks~\cite{Bai2022TowardsModels}  or by learning pruning masks for block weights~\cite{Xu2024BESA:Allocation}. In contrast, we derive an upper bound on the reconstruction error and introduce a pruning metric to minimize this bound. This enables highly efficient pruning in a single forward pass, eliminating the computational cost associated with training blocks or learning masks.

\vspace{-0.5em}

\section{Method}
\label{sec:Method}
\subsection{Preliminary}


\noindent \textbf{A transformer block} $f(\mathbf{X}) $ with input $\mathbf{X}$ consists of a
multi-head self-attention (MSA) module and a Feed-Forward Network (FFN) module:
\begin{align}
\label{MSA module}
\mathbf{h}_{i}(\mathbf{X}) &= \text{softmax}\left(\frac{\mathbf{Q}_{i}{\mathbf{K}_{i}}^T}{\sqrt{d}}\right)\mathbf{V}_{i},\\
\text{MSA}(\mathbf{X}) &= \text{Concat}\left[\mathbf{h}_{1},\ldots,\mathbf{h}_{n}\right]\mathbf{W}^{O}, \\
\mathbf{X}^{'}&=\text{MSA}(\mathbf{X})+\mathbf{X},\\
\label{MLP module}
\text{FFN}(\mathbf{X}^{'}) &=\sigma(\mathbf{X}^{'}\mathbf{W}^{U})\mathbf{W}^{D},\\
\label{transformer block function}
f(\mathbf{X}) &= \text{FFN}(\mathbf{X}^{'})+\mathbf{X}^{'},
\end{align}
where $\mathbf{X}$ and $\mathbf{X}^{'}$ are the input to MSA and FFN modules respectively. $\mathbf{Q}_{i}$, $\mathbf{K}_{i}$, and $\mathbf{V}_{i}$ denote the query, key, and value of the $i$-th attention head $\mathbf{h}_{i}$. $\mathbf{W}^{U}$, $\mathbf{W}^{D}$ and $\mathbf{W}^{O}$ are the weight matrices of FFN up projection layer, FFN down projection layer and MSA output projection layer, respectively. $\sigma$ is an activation function. $d$ is the embedding dimension of input tokens, and $n$ is the number of attention heads in MSA.

\noindent \textbf{Dependency-aware structured pruning.} In structured pruning, connected components should be pruned together to maintain the model's integrity and functionality. For the FFN module, we prune channels in the FFN hidden layer based on their channel importance scores $\mathbf{s}^F$, where $\mathbf{s}$ denotes a vector of importance scores and superscript $F$ denotes channels in the FFN hidden layer. The connected weights in FFN up and gate projection layers are pruned accordingly. For the MSA module, we prune the attention heads together with the connected channels in query, key, value, and output projection layers. To obtain the importance score of one attention head, we sum up the importance scores $\mathbf{s}^H$ of corresponding output channels from this head,
where superscript $H$ denotes output channels of attention heads. We then prune the attention heads with low attention head scores.
In the subsequent section, we elaborate on the calculation of channel importance scores $\mathbf{s}^{H}$ and $\mathbf{s}^F$.

\subsection{Problem Definition}
 Our goal is to minimize the impact of the pruned channels on the transformer block output. Mathematically, it aims to predict two binary masks $\overline{\mathbf{s}}^H$ (0 for pruning and 1 for keeping) and $\overline{\mathbf{s}}^F$ such that the reconstruction error $\xi$  between the output $f(\mathbf{X})$ of the original unpruned transformer block and the output $f(\mathbf{X},\overline{\mathbf{s}}^H,\overline{\mathbf{s}}^F)$ produced by the pruned block is minimized under the constraint of target sparsity ratio $r$:
 
\begin{equation}
\begin{aligned}
\label{opt obj}
\min_{\overline{\mathbf{s}}^H,\overline{\mathbf{s}}^F} \xi &= | f(\mathbf{X}) - f(\mathbf{X},\overline{\mathbf{s}}^H,\overline{\mathbf{s}}^F)|, \\
\text{s.t.} \left\|\overline{\mathbf{s}}^H \right\|_0 &= (1-r)*N^{H}, \overline{\mathbf{s}}^H \in \{0,1\}^{N^{H}}, \\
\left\| \overline{\mathbf{s}}^F \right\|_0 &= (1-r)*N^{F}, \overline{\mathbf{s}}^F \in \{0,1\}^{N^{F}}
\end{aligned}
\end{equation}
where $N^{H}$ and $N^{F}$ are the number of channels for attention heads and FFN hidden layer respectively. $\left\|\cdot\right\|_0$ is the number of non-zero elements. The reconstruction error $\xi$ is defined as the element-wise absolute value ($|\cdot|$).
This choice is primarily for the convenience of  subsequent derivations.

The optimal solution for Eq.~(\ref{opt obj}) requires a joint search of $\mathbf{\overline{s}}^{H}$ and $\mathbf{\overline{s}}^F$, which is computationally costly. To reduce computation cost, we optimize $\mathbf{\overline{s}}^H$ and $\mathbf{\overline{s}}^F$ separately.

\subsection{Pruning Metric}

Directly optimizing Eq.~(\ref{opt obj}) is still computationally expensive. This is because there are often several thousands of channels in LLMs, resulting in a large search space. Additionally, for each searched $\mathbf{\overline{s}}$, one forward pass through the transformer block is needed to evaluate the solution. To reduce computation cost, we derive an upper bound of this objective in Eq.~(\ref{opt obj}) and minimize it with only \textit{one forward pass} through the transformer block.  From Eqs.~(\ref{MLP module}), (\ref{transformer block function}) and (\ref{opt obj}), we have:

\begin{equation}
\label{sl2 derivation}
\begin{aligned}
&| f(\mathbf{X}) - f(\mathbf{X},\overline{\mathbf{s}}^F)|\\
&= |(\text{FFN}(\mathbf{X}^{'})+\mathbf{X}^{'})-(\text{FFN}(\mathbf{X}^{'},\overline{\mathbf{s}}^F)+\mathbf{X}^{'})|\\
&= |\sigma(\mathbf{X}^{'}\mathbf{W}^{U})\mathbf{W}^{D}-\sigma(\overline{\mathbf{s}}^F \odot \mathbf{X}^{'}\mathbf{W}^{U})\mathbf{W}^{D}|\\
&\le |\sigma(\mathbf{X}^{U})-\sigma(\overline{\mathbf{s}}^F \odot \mathbf{X}^{U})|\cdot|\mathbf{W}^{D}|\\
&\le C_{\sigma}|\mathbf{X}^{U}-\overline{\mathbf{s}}^F \odot \mathbf{X}^{U}|\cdot|\mathbf{W}^{D}|\\
&= C_{\sigma}|1-\mathbf{\overline{s}}^F| \odot |\mathbf{X}^{U}|\cdot|\mathbf{W}^{D}|\\
&= C_{\sigma}\sum_{j=1}^{N^F}(1-\mathbf{\overline{s}}_{j}^F) |\mathbf{X}_{j}^{U}|\cdot|\mathbf{W}_j^{D}|,
\end{aligned}
\end{equation}
where $\cdot$ is dot product and $\odot$ is the element-wise product. $\mathbf{X}_{j}^{U}$ and $\mathbf{W}_{j}^{D}$ are the output value and output weights of $j$-th channel in the FFN hidden layer.  In this derivation, the activation function $\sigma$ is assumed to be Lipschitz continuous with the Lipschitz constant of $C_{\sigma}$, and this holds for most activation functions (e.g., \rm{GeLU}, ReLU, Swish) in LLMs. 

The bottom line in Eq.~(\ref{sl2 derivation}) is the upper bound for the objective in Eq.~(\ref{opt obj}). To minimize this upper bound, $\mathbf{\overline{s}}_{j}^F$ should be set to one for the channels with large $|\mathbf{X}_{j}^{U}|\cdot|\mathbf{W}_j^{D}|$ and zero for the channel with small values. In other words, channels with small $|\mathbf{X}_{j}^{U}|\cdot|\mathbf{W}_j^{D}|$ should be pruned. This means that $|\mathbf{X}_{j}^{U}|\cdot|\mathbf{W}_j^{D}|$ serves as the importance score for the $j$-th channel in the FFN hidden layer. Similarly, we can derive that $|\mathbf{X}_{j}^{H}|\cdot|\mathbf{W}_j^{O}|(I+|\mathbf{W}^{U}||\mathbf{W}^{D}|)$ is the importance score for the $j$-th output channel of the MSA attention heads (as detailed in the supplementary materials). Based on the derivations, we define the channel importance of a transformer block as follows:

\noindent  \textbf{Definition 1. (Block-wise importance scores)} Given a transformer block, the importance score $ \mathbf{s}_{j}^H$ for the $j$-th output channel of MSA attention heads and $ \mathbf{s}_{j}^F$ for  the $j$-th channel of  FFN hidden layer in this block are:
\begin{align}
\label{eq:prune_metric_MSA}
\text{MSA:} \quad \mathbf{s}_{j}^H&=  |\mathbf{X}_{j}^{H}|\cdot|\mathbf{W}_j^{O}|(\mathbf{I}+|\mathbf{W}^{U}||\mathbf{W}^{D}|),\\
\label{eq:prune_metric_MLP}
\text{FFN:}\quad   \mathbf{s}_{j}^F&= |\mathbf{X}_{j}^{U}|\cdot|\mathbf{W}_j^{D}|,
\end{align}
where $\mathbf{X}_{j}^{U}$ and $\mathbf{W}_{j}^{D}$ are the output value and output weights of the $j$-th channel in the FFN hidden layer. $\mathbf{X}_{j}^{H}$ and $\mathbf{W}_{j}^{O}$ are output value and output weights of the $j$-th channel from the MSA attention heads. The entire procedure is illustrated in Algorithm~\ref{alg:overview}.

\begin{algorithm}[tb]
\caption{Block-Wise Pruning for LLMs}
\label{alg:overview}
\begin{algorithmic}
\STATE {\bfseries Input:} Pre-trained model $M$, number of transformer blocks $L$, target sparsity ratio $r$, calibration set $X$
\FOR{$l = 1$ to $L$}
\STATE \textbf{Access} weight matrices ($\mathbf{W}^{O}$, $\mathbf{W}^{U}$, $\mathbf{W}^{D}$) of the $l$-th transformer block
\STATE \textbf{Obtain} the output  $\mathbf{X}^{H}$ of the attention heads, and output $\mathbf{X}^{U}$  of the FFN hidden layer via forward propagation
\STATE \textbf{Prune} attention heads by aggregating the head-wise channel importance scores in Eq.~(\ref{eq:prune_metric_MSA}), and prune the connected channels in the query, key, value, and projection layers.
\STATE \textbf{Prune} channels of the FFN hidden layer using Eq.~(\ref{eq:prune_metric_MLP}), and the connected weights in FFN up and gate projection layers to the target sparsity level $r$
\ENDFOR
\STATE {\bfseries Output:} Pruned model $\overline{M}$
\end{algorithmic}
\end{algorithm}

\noindent \textbf{Remark.} The proposed pruning approach offers multiple advantages over existing works in terms of accuracy, computational efficiency, and memory efficiency. Compared to gradient-based global pruning approaches \cite{Ma2023LLM-Pruner:Models, Zhang2023LoRAPrune:Fine-Tuning}, our method is more accurate and efficient because we only need one-shot forward pass and it does not require computing and storing near-zero and likely unreliable gradients for pruning decisions. Unlike importance backpropagation \citep{Yu2018NISP:Propagation}, which relies on the Lipschitz continuous assumption of the self-attention module to globally prune channels, our method is accurate because it does not rely on this invalid Lipschitz continuous assumption \cite{Kim2021TheSelf-Attention}, instead pruning channels locally within the transformer block. Compared to Wanda \citep{Sun2024AModels}, which targets the minimization of prune-induced error in a single layer (e.g., projection layer, Query, Key, Value), our method is more optimal as it targets the minimization of reconstruction error in the transformer block output, thus reducing the issue of error accumulation.

\section{Experiments}
\label{sec:Experiments}

\begin{table*}[t]
    \centering
    \caption{Zero-shot performance of the compressed LLaMA-7B model. 
    The average is calculated among seven classification datasets. 
    } \label{tab:LLaMA-7b_result}
    \vspace{-1em}
    \resizebox{\linewidth}{!}{
    \begin{tabular}{ll|cc|cccccccc}
        \toprule
        \toprule
        Pruning Ratio & Method & WikiText2$\downarrow$ & PTB$\downarrow$ & BoolQ & PIQA & HellaSwag & WinoGrande & ARC-e & ARC-c & OBQA & Average$\uparrow$ \\
        \midrule  
        \multirow{1}{*}{Ratio = 0\%} & LLaMA-7B~\citep{Touvron2023LLaMA:Models} & 5.69 & 8.93 & 76.50 & 79.80 & 76.10 & 70.10 & 72.80 & 47.60 & 57.20 & 68.59 \\
        \cmidrule{1-12}
        \cmidrule{1-12}
        \multirow{6}{*}{\parbox{1.9cm}{Ratio = 20\% \  w/o tune}} 
        & Random & 27.51 & 43.19 & 61.83 & 71.33 & 56.26 & 54.46 & 57.07 & 32.85 & 35.00 & 52.69\\
        &Magnitude& 26.98 & 44.78 & 61.93 & 69.89 & 58.87 & 55.12 & 56.93 & 32.54 & 36.10 & 53.05 \\
        &Wanda\citep{Sun2024AModels} & 22.12 & 38.19 & 64.93 & 70.14 & 58.12 & 55.39 & 56.63 & 33.98 & 35.43 & 53.23\\
        & LLM-Pruner\citep{Ma2023LLM-Pruner:Models}& 19.77 & \textbf{36.66} & 59.39 & 75.57 & 65.34 & 61.33 & 59.18 & 37.12 & 39.80 & 56.82 \\
        & Importace Propagation\citep{Yu2018NISP:Propagation} &23.39 & 37.16& 65.47& 67.08& 37.63&56.82 & 51.22 &26.70 & 21.00 & 46.56\\
        \rowcolor{myblue}& LLM-BIP & \textbf{19.63} & 37.45 & \textbf{71.13} & \textbf{77.26} & \textbf{73.13} & \textbf{66.22} & \textbf{66.67} & \textbf{40.78} & \textbf{42.8} & \textbf{62.57} \\
        \cmidrule{1-12}
        \multirow{4}{*}{\parbox{1.9cm}{Ratio = 20\% \ w/ tune}} 
        &Magnitude& 21.78 & 38.64 & 61.89 & 70.81 & 58.34 & 56.87 & 54.87 & 34.02 & 38.40 & 53.59 \\
        &Wanda& 18.43 & 33.16 & 65.75 & 74.70 & 64.52 & 59.35 & 60.65 & 36.26 & 39.40 & 57.23 \\
        & LLM-Pruner & 17.58 &  30.11 & 64.62 & 77.20 &  68.80 & 63.14 & 64.31 & 36.77 & 39.80 & 59.23 \\
        \rowcolor{myblue}& LLM-BIP & \textbf{16.98} & \textbf{29.86} & \textbf{72.02} & 77.85 & \textbf{71.29} & \textbf{66.92} & 64.98 & \textbf{39.84} & \textbf{43.6} & \textbf{62.36}\\
        \cmidrule{1-12}
        \multirow{6}{*}{\parbox{1.9cm}{Ratio = 50\% w/o tune}} 
        & Random& 3887.90 & 4337.27 & 46.79 & 53.37 & 27.50 & 50.59 & 28.07 & 27.90 & 30.00 & 37.75 \\
        &Magnitude& 785.10 & 1588.86 & 44.10 & 54.98 & 31.27 & 52.93 & 38.76 & 27.50 & 29.67 & 39.88 \\
        &Wanda& 223.46 & 437.71 & 45.13 & 55.54 & 31.37 & 55.87 & 39.43 & 25.76 & 30.12 & 40.46 \\
        & LLM-Pruner&112.44 & 255.38 & 52.32 & 59.63 & 35.64 & 53.20 & 33.50 & 27.22 & \textbf{33.40} & 42.13 \\
        & Importace Propagation &4042.79 &6717.69 & 49.69& 52.07 & 28.87 & 51.46 & 28.07 & 25.85 &25.20 &37.32 \\
        \rowcolor{myblue} & LLM-BIP & \textbf{72.00} & \textbf{109.86} & \textbf{53.12} & \textbf{65.07} & \textbf{44.37} & \textbf{55.25} & \textbf{44.57} & \textbf{28.50} & 32.40 & \textbf{46.18}\\
          \cmidrule{1-12}
         \multirow{5}{*}{\parbox{1.9cm}{Ratio = 50\% w/ tune}}
         &Magnitude& 78.80 & 164.32 &  47.40 & 54.36 & 33.49 & 53.10 & 37.88 & 26.60 & 30.12 & 40.42 \\
        &Wanda& 43.89 & 85.87 &  50.90 & 57.38 & 38.12 & 55.98 & 42.68 & \textbf{34.20} & \textbf{38.78} & 45.43 \\
        & LLM-Pruner& 38.12 & 66.35 & 60.28 & 69.31 & 47.06 & 53.43 & \textbf{45.96} & 29.18 & 35.60 & \textbf{48.69} \\
        & Importance Propagation & 59.50 & 87.93 & 56.73 & 60.55 &38.47 &50.59  & 36.36 & 23.81 & 29.00 & 42.22 \\        
        \rowcolor{myblue} & LLM-BIP & \textbf{32.67} & \textbf{46.88}  & 51.38 & 67.68 & \textbf{49.81} & \textbf{56.70} & 45.62 & 28.42 & 31.20 & 47.26 \\
        \bottomrule
        \bottomrule
    \end{tabular}
    }
\end{table*}

\begin{table*}[t]
    \centering
    \caption{Zero-shot performance of the compressed Vicuna-7B model. 
    The average is calculated among seven classification datasets. 
    } \label{tab:Vicuna-7B_result}
    \vspace{-1em}
    \resizebox{\linewidth}{!}{
    \begin{tabular}{ll|cc|cccccccc}
        \toprule
        \toprule
        Pruning Ratio & Method & WikiText2$\downarrow$ & PTB$\downarrow$ & BoolQ & PIQA & HellaSwag & WinoGrande & ARC-e & ARC-c & OBQA & Average$\uparrow$ \\
        \midrule  
        \multirow{1}{*}{Ratio = 0\%}
         & Vicuna-7B&16.11  &61.37 &76.57 & 77.75 & 70.64 & 67.40 & 65.11 & 41.21& 40.80 & 62.78\\
        \cmidrule{1-12}
        \cmidrule{1-12}
        \multirow{6}{*}{\parbox{1.9cm}{Ratio = 20\% \  w/o tune}} 
        & Random & 36.02 & 106.90 &61.47 & 70.89 &54.67 & 56.27 & 55.60 & 31.74 & 34.60 & 52.18 \\
        &Magnitude & 2189.45 & 2549.75 & 55.90 & 56.15 & 32.37 & 51.85 & 30.01 & 28.41 &28.20 &40.41 \\
        & LLM-Pruner&\textbf{25.74} &92.88 &61.70 &75.30 & 63.75 & 56.20 & 63.22 & 36.60 & 37.00 & 56.25\\
        & Importace Propagation& 88.97 & 202.86 &63.02 &62.13 & 36.17& 57.46 & 50.29& 28.84 & 19.40 & 45.33\\
        &Wanda & 47.53 & 144.41 & 55.75&71.16 & 44.39& 57.06 & 64.60 & 34.47 & 26.60 & 50.58\\
        \rowcolor{myblue} & LLM-BIP & 27.08 & \textbf{92.73} & \textbf{72.54} & \textbf{75.84} & \textbf{71.13} &\textbf{64.16}  & \textbf{65.53} & \textbf{42.24} & \textbf{41.60} & \textbf{61.86}  \\
                \midrule
        \cmidrule{1-12}
        \multirow{4}{*}{\parbox{1.9cm}{Ratio = 20\% \ w/tune}} 
        & LLM-Pruner &19.69 & 78.25 & 63.33 & 76.17 & 65.13 & 60.22 & 62.84 & 37.12 & 39.20 & 57.71\\
        & Importace Propagation & 88.97 & 202.86 & 65.47 &69.31 &54.87 &59.51 & 53.24& 30.89& 33.20& 52.36\\
        &Wanda& 47.53 & 144.41 & 68.20 & 75.46 & 66.08 & 64.01 & 65.24 & 37.80 & 40.00 & 59.54 \\
       \rowcolor{myblue} & LLM-BIP & \textbf{18.15} & \textbf{60.20} & \textbf{72.02} & \textbf{76.61} & \textbf{70.29} & \textbf{64.80} & \textbf{65.87} & \textbf{39.68} & \textbf{40.20} &  \textbf{61.35} \\
        \cmidrule{1-12}
        \multirow{2}{*}{\parbox{1.9cm}{Ratio = 50\% \  w/o tune}} 
         & LLM-Pruner & 143.85 & 427.77 & 53.76 & 59.79 & 34.86 & \textbf{50.28} & 33.29 & 27.30 & \textbf{34.60} & 41.98 \\
        &Wanda& 331.16 & 518.72	 & \textbf{56.61} & 57.94 & 35.34 & 49.88 & 37.46 & \textbf{26.54} & 26.80 & 41.51 \\
       \rowcolor{myblue} & LLM-BIP & \textbf{80.38} & \textbf{189.82} & 51.74 & \textbf{61.10} & \textbf{36.28} & 49.88 & \textbf{41.03} & 24.43 & 29.60 &  \textbf{42.01} \\
        \bottomrule
        \bottomrule
    \end{tabular}
    }
    \vspace{-1em}
\end{table*}

\begin{table*}[t]
    \centering
    \caption{Zero-shot performance of the compressed LLaMA-13B models. 
    The average is calculated among seven classification datasets. 
    } \label{tab:LLaMA-13b_result}
    \vspace{-1em}
    \resizebox{\linewidth}{!}{
    \begin{tabular}{ll|cc|cccccccc}
        \toprule
        \toprule
        Pruning Ratio & Method & WikiText2$\downarrow$ & PTB$\downarrow$ & BoolQ & PIQA & HellaSwag & WinoGrande & ARC-e & ARC-c & OBQA & Average$\uparrow$ \\
        \midrule  
        \multirow{1}{*}{Ratio = 0\%}
         & LLaMA-13B\citep{Ma2023LLM-Pruner:Models} & 11.58 & 20.24 &  68.47 & 78.89 & 76.24 &  70.09 & 74.58 & 44.54 & 42.00  & 64.97 \\
        \cmidrule{1-12}
        \cmidrule{1-12}
        \multirow{6}{*}{\parbox{1.9cm}{Ratio = 20\% \  w/o tune}} 
        & Random& 20.56 &91.43  & 63.33 & 73.18 & 63.54 &60.85 &64.44 &36.26 &38.00 & 57.09\\
        &Magnitude& 471.66 & 742.02 &61.50 & 67.57& 52.90 & 57.54&50.13 & 31.14&36.80 &51.08\\
        & LLM-Pruner& 17.05 & 83.25 & 67.68 &  77.15 & 73.41 & 65.11 & 68.35 & 38.40 & 42.40 & 61.79 \\
        & Importace Propagation& 60.20 & 141.62 & 58.72 & 68.12 & 41.37 & 62.35 &54.67 & 30.97 & 21.40 & 48.23 \\
        &Wanda & 20.93 & 91.08 &74.22 & 77.53 & 54.14 & 67.32 & 70.24 & 38.73 &32.40 & 59.23 \\
       \rowcolor{myblue} & LLM-BIP & \textbf{16.07} & \textbf{74.01} & \textbf{80.00} & \textbf{79.33} & \textbf{77.02} & \textbf{71.90} & \textbf{73.95} & \textbf{47.01} & \textbf{44.80} & \textbf{67.72} \\
        \cmidrule{1-12} 
         \multirow{2}{*}{\parbox{1.9cm}{Ratio = 50\% \  w/o tune}} 
        & LLM-Pruner& 73.18 & 387.97 & 38.90 & 65.34 & 45.05 & 53.20 & 36.53 & 27.82 & \textbf{35.40} & 43.18 \\
        &Wanda & 128.94 & 364.47 & 57.95 & 58.71 & 43.41 & 53.04 & 41.58 & 27.22 & 31.00 &  44.70\\
       \rowcolor{myblue}  & LLM-BIP & \textbf{57.67} & \textbf{183.27} & \textbf{64.40} & \textbf{69.21} & \textbf{52.67} & \textbf{57.30} & \textbf{46.46} & \textbf{30.12} & 35.20 & \textbf{	50.77} \\
               \cmidrule{1-12} 
         \multirow{2}{*}{\parbox{1.9cm}{Ratio = 70\% \  w/o tune}} 
        & LLM-Pruner & 605.62 & 1075.42  & 45.23 &  \textbf{57.83} & 28.90 & \textbf{52.64} & 28.66 & 24.74 & 25.60 & 37.66 \\
        &Wanda & 384.95 & 694.34 & 55.11 & 55.93 & 30.01 & 50.20 & 29.76 & 22.44 & 24.80 & 38.32 \\
       \rowcolor{myblue}  & LLM-BIP & \textbf{157.98} & \textbf{349.14} & \textbf{61.35} & 54.68 & \textbf{30.62} & 48.70 & \textbf{30.26} & \textbf{24.15} & \textbf{28.60} & \textbf{39.77} \\
        \bottomrule
        \bottomrule
    \end{tabular}
    }
\end{table*}

\subsection{Experimental Settings}
\textbf{LLMs.} To demonstrate the effectiveness and versatility of the proposed method,  we evaluate it on three open-source large language models with varied model sizes: LLaMA-1-7B, LLaMA-2-13B \citep{Touvron2023LLaMA:Models}, and Vicuna-7B \citep{Wei-LinChiang2023Vicuna:Quality}.

\noindent \textbf{Evaluation and datasets.} We adopt LLaMA's evaluation protocol to conduct zero-shot task classification on common-sense reasoning datasets: BoolQ~\cite{Clark2019BoolQ:Questions}, PIQA~\cite{Bisk2020PIQA:Language}, HellaSwag~\cite{Zellers2019HellaSwag:Sentence}, WinoGrande~\cite{Sakaguchi2019WinoGrande:Scale}, ARC-easy~\cite{Clark2018ThinkChallenge}, ARC-challenge~\cite{Clark2018ThinkChallenge} and OpenbookQA~\cite{Mihaylov2018CanAnswering}. Additional evaluation is conducted with a zero-shot perplexity (PPL) analysis on WikiText2~\cite{Merity2017PointerModels} and PTB~\cite{Marcus1993BuildingTreebank}.

\noindent  \textbf{Implementation details.} During the model pruning phase, 128 sentences from C4~\cite{Raffel2019ExploringTransformer} are randomly sampled to perform the one-shot pruning, and each of these sentences is truncated to a sequence length of 2048. After pruning, the model undergoes finetuning using LoRA~\cite{Hu2022LoRA:Models}, the low-rank approximation technique. For finetuning, 20,000 sentences with a length of 512 tokens from C4~\cite{Raffel2019ExploringTransformer} are used to fine-tune the pruned model. The model is fine-tuned using the AdamW optimizer~\cite{Loshchilov2019DecoupledRegularization} on a single A100 GPU (80G) for 1 epoch with an initial learning rate of 0.0001 and a batch size of 128. Cosine learning rate decay~\cite{Loshchilov2017SGDR:Restarts} is employed.

\noindent  \textbf{Contenders.} We compare the proposed method with the following baselines: \textbf{1)} Random pruning that removes channels randomly; \textbf{2)} Magnitude-based pruning that removes channels with the smallest sum of absolute values of connected weights; \textbf{3)} LLM-Pruner~\cite{Ma2023LLM-Pruner:Models}: evaluating channel importance by aggregating first/second-order  gradients of connected weights; \textbf{4)} Importance backpropagation~\cite{Yu2018NISP:Propagation}: propagating the importance score of the final layer to the early layers for evaluating channel importance; \textbf{5)} Wanda~\cite{Sun2024AModels}: pruning channels with the smallest impact on the current layer's output.

\subsection{Zero-shot Performance} Tables ~\ref{tab:LLaMA-7b_result}, \ref{tab:Vicuna-7B_result} and \ref{tab:LLaMA-13b_result} present the zero-shot performance of the pruned model. Evaluation on LLaMA-7B, Vicuna-7B and LLaMA-13B models demonstrates that with a 20\% reduction in parameters without fine-tuning, the proposed method surpasses LLM-Pruner (the strongest baseline) by 1.40 on average on  WikiText2 and PTB datasets. The robust performance of the proposed method is further confirmed by the common-sense reasoning tasks where the proposed method achieves 5.76\% higher accuracy than LLM-Pruner, while maintaining 94.88\% performance of the unpruned model.  With a higher sparsity level (50\% parameters pruned), our method outperforms the baselines by a much more significant margin, achieving 117.93 lower perplexities on WikiText2 and PTB datasets and around 3.89\% higher accuracy on common reasoning tasks compared to the best contender. The remarkable performance of our method at high sparsity levels should be attributed to the more accurate channel pruning employed by our approach. 

We compare the fine-tuned performance of our pruned models with baselines. As depicted in Tables~\ref{tab:LLaMA-7b_result} and \ref{tab:Vicuna-7B_result}, we achieve similar or higher accuracies compared to the baselines after fine-tuning, highlighting the necessity of fine-tuning to enhance accuracy.
Specifically, our fine-tuned pruned models yield approximately 2.44\% higher accuracy on seven common reasoning tasks and reduce an average of 7.56 perplexities on WikiText2 and PTB datasets compared to LLM-pruner. 

Notably, fine-tuning offers no significant accuracy improvement for models pruned by our method on common reasoning tasks at the sparsity of 20\% for both LLaMa-7B and Vicuna-7B. This suggests that our pruned models may already be near the optimal accuracy achievable through fine-tuning on common reasoning tasks. This is supported by a previous study~\cite{Zhang2023LoRAPrune:Fine-Tuning}, where fine-tuning a 20\% sparsity LLaMA-7B model on the larger LaMini-instruction dataset~\cite{Wu2024LaMini-LM:Instructions} with 2.58 million samples resulted in an accuracy of 62.70\%, closely matching our pruned model's 62.57\% accuracy without fine-tuning. However, fine-tuning remains necessary at higher sparsity levels (e.g., 50\%) to improve performance, as our 50\% sparsity LLaMA-7B model slightly underperforms LLM-pruner on common reasoning tasks after fine-tuning  due to the smaller fine-tuning dataset (20k samples vs 50k samples) and fewer fine-tuning epochs (1 epoch vs 2 epochs).

Our method outperforms the baselines due to a more appropriate ``pruning scope". In contrast to global pruning approaches, our method avoids the invalid Lipschitz continuous assumption on the self-attention module~\cite{Kim2021TheSelf-Attention} and the dependence on near-zero gradients for pruning decisions. Unlike the layerwise pruning approach, we minimize the pruning impact within the scope of a transformer block rather than a single layer, resulting in reduced error accumulation and improved efficacy. 

\begin{figure*}[t]
  \centering
    \centering
    \includegraphics[width=0.95\linewidth,height=2.35in]{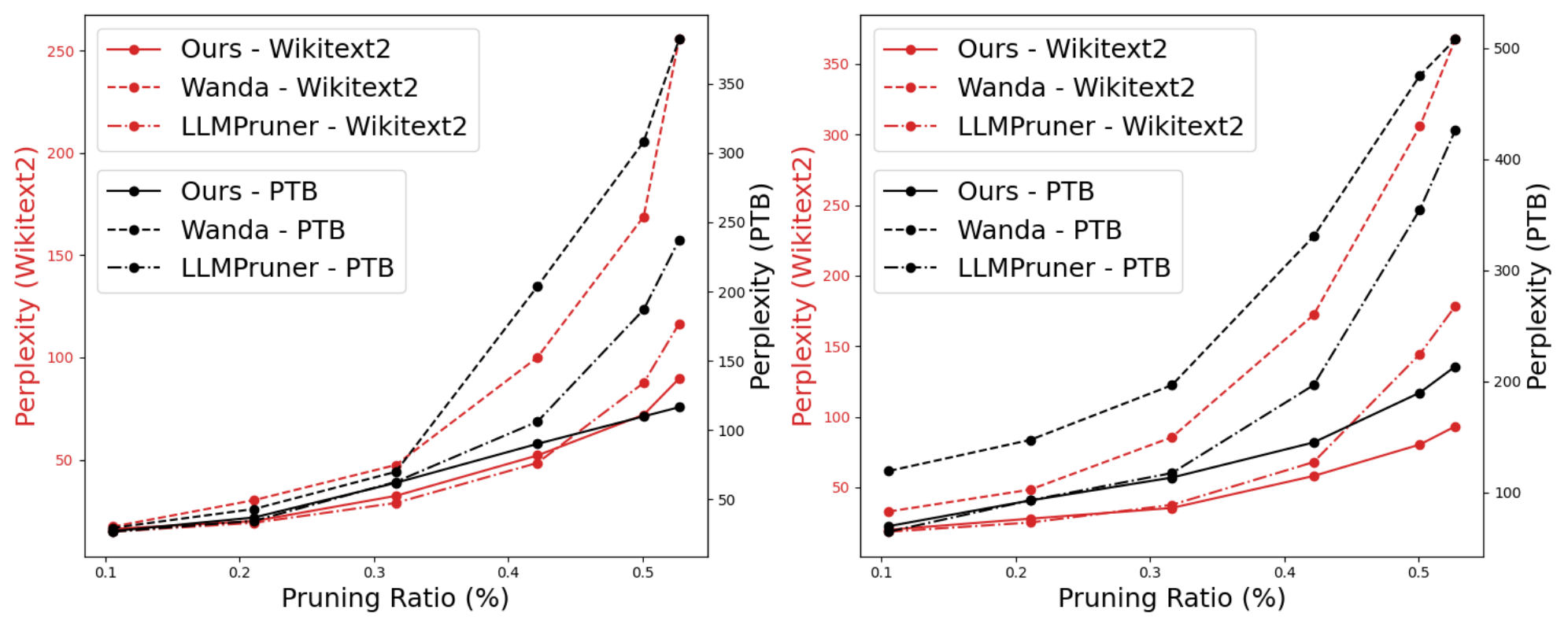}
    \caption{Comparison of the proposed method, Wanda and LLM-Pruner on LLaMA-7B (left) and Vicuna-7B (right) with different pruning rates without fine-tuning.}
    \label{fig:pruning_ratio}
\end{figure*}

\subsection{Ablation Study}

\begin{table}[t]
\caption{Comparison of the pruning speed (seconds). We use 128 sentences with a sequence length of 512 from C4. Speed is measured on a NVIDIA A100 GPU.}
\label{analysis: pruning efficiency}
\centering
  \setlength{\tabcolsep}{2pt}
  \resizebox{\linewidth}{!}{
\begin{tabular}{@{} *{1}{c} *{1}{c} *{1}{c} *{1}{c} @{}}
\toprule
Method & LLaMA-7B
& Vicuna-7B
& LLaMA-13B\\
\midrule
 LLM-Pruner~\cite{Ma2023LLM-Pruner:Models} & 34.22 & 34.18 & 61.04 \\
Wanda~\cite{Sun2024AModels} & 16.01 & 16.05 & 26.87 \\
 LLM-BIP & \textbf{15.76} & \textbf{15.78}
 & \textbf{26.50}   \\
\bottomrule
\end{tabular}
}
\end{table}

\begin{table}
\caption{Comparison between the original model and the compressed model. 128 sentences with a sequence length of 512 from C4 are used. The metrics are the parameter count (\#Params), multiply-accumulate operations (MACs), and inference latency (seconds).  Inference latency is measured on a NVIDIA A100 GPU. }\label{tbl:stat_param}
 \resizebox{\linewidth}{!}{
    \begin{tabular}{c|c|ccc}
        \toprule
        Model  & Pruning ratio & \#Params & \#MACs & Latency \\
        \midrule
         LLaMA-7B & 0\% & 6.74B & 3451.99G  & 8.60s \\
         LLaMA-7B & 20\% & 5.37B & 2737.99G  & 6.53s\\
          Vicuna-7B & 20\% & 5.37B &  2737.99G  & 6.67s\\
         LLaMA-7B & 50\% & 3.49B & 1756.24G  & 5.17s \\
        \bottomrule
    \end{tabular}   
}
\end{table}

\noindent \textbf{Impact of different pruning rates.} We compare our method with Wanda and LLM-pruner across various pruning ratios. Figure~\ref{fig:pruning_ratio} illustrates that Wanda exhibits a rapid increase in perplexity at approximately 30\% sparsity. Similarly, LLM-pruner demonstrates a significant perplexity increase when sparsity is increased from 20\% to 50\% on the LLaMA-7B and Vicuna-7B models. In contrast, our proposed method exhibits greater robustness to changes in sparsity.

\noindent \textbf{Pruning efficiency.}  We empirically compare our method with LLM-Pruner and Wanda pruning speed. Table~\ref{analysis: pruning efficiency} shows both Wanda and our method are 2 times faster than LLM-Pruner. This is because Wanda and our method only need one forward process to prune the channel while LLM-Pruner needs a forward and a backward processes to compute and store the gradients for pruning decisions.

\begin{figure}[t]
\begin{center}
\includegraphics[width=0.8\linewidth]{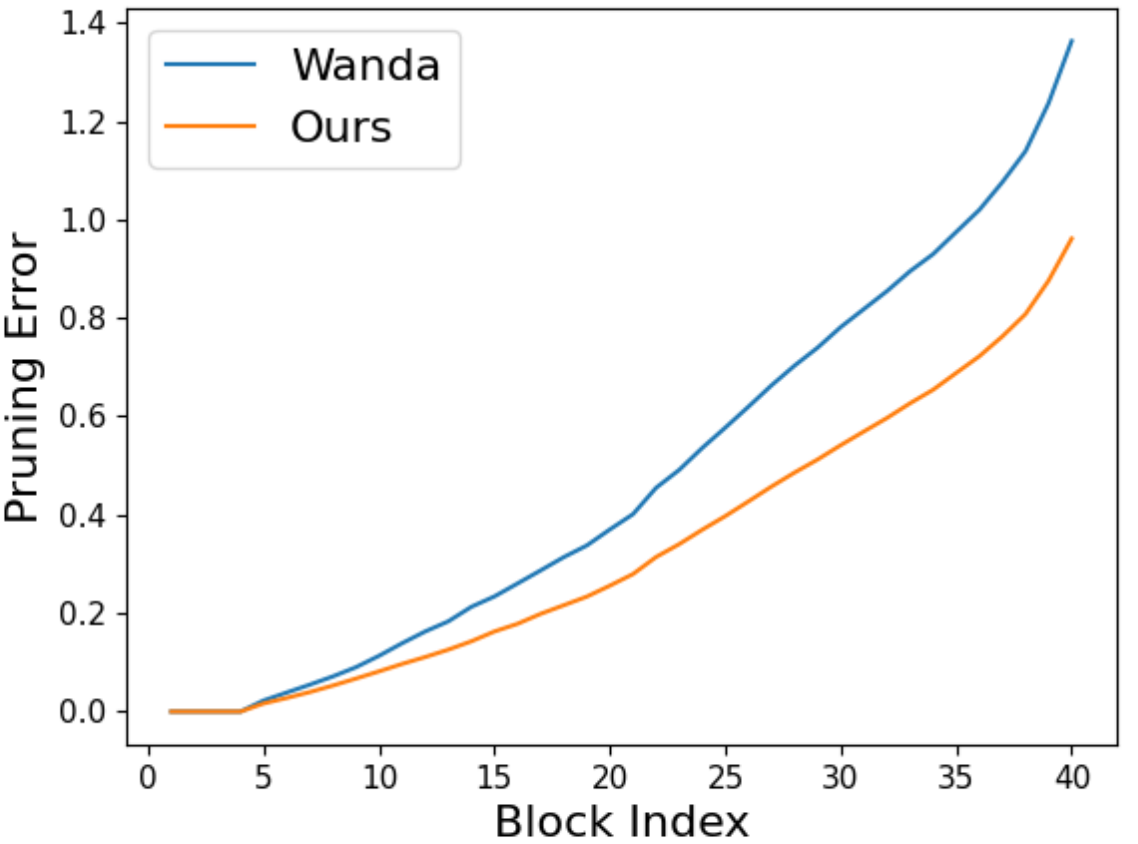}
\end{center}
   \caption{Our method mitigates the pruning error accumulation issue compared to Wanda. Pruning error is measured by the mean value of the distance defined by Eq.~(\ref{opt obj}) for each transformer block output. The LLaMA-13B model is pruned to the sparsity of 20\% without finetuing. }
\label{fig: pruning_error}
\end{figure}

\begin{figure}[t]
\begin{center}
\includegraphics[width=1.0\linewidth]{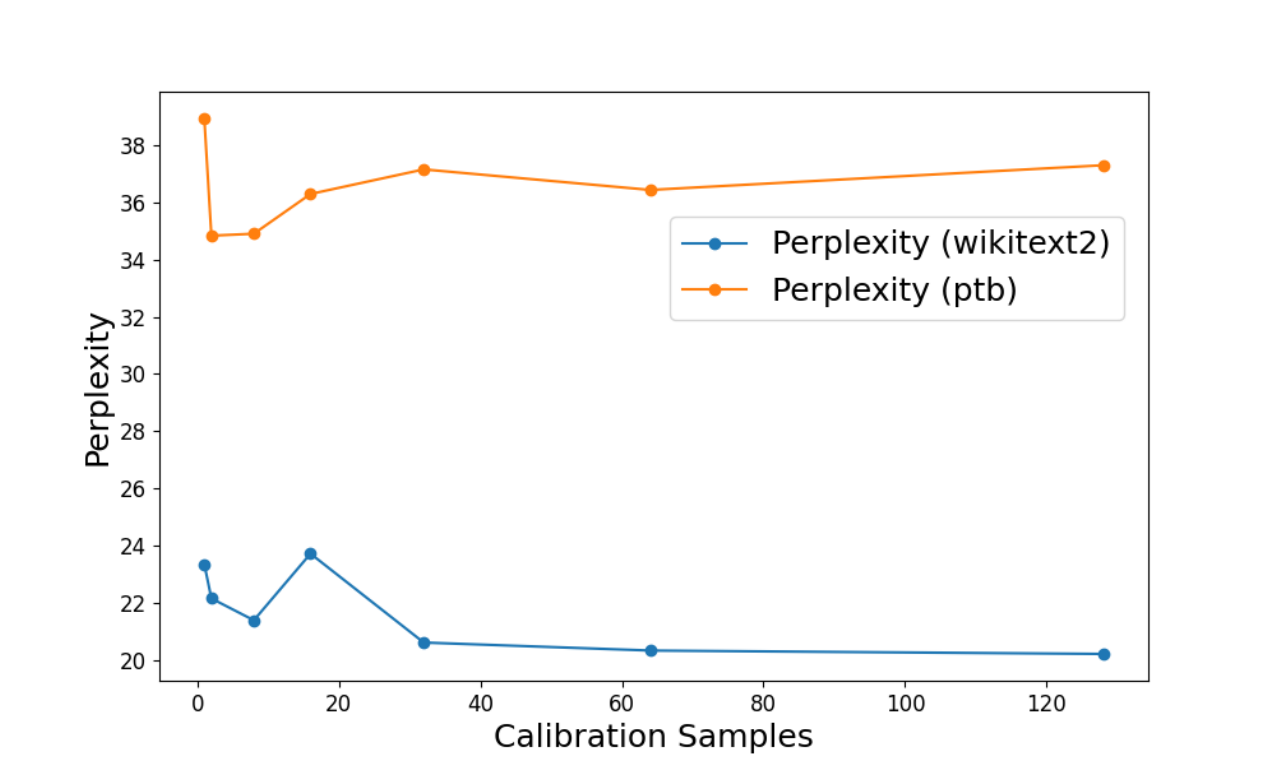}
\end{center}
   \caption{Our method is robust to the calibration set size. The calibration set consists of sentences with a sequence length of 2048 from C4. The LLaMA-7B model is pruned to the sparsity of 20\% without finetuing. }
\label{fig: calibration_set_size}
\end{figure}

\noindent \textbf{Inference speed.} We investigate the inference acceleration achieved by our proposed method. Table~\ref{tbl:stat_param} demonstrates that, owing to structural pruning, our method reduces inference latency time by 24\% and 40\% at compression ratios of 20\% and 50\%, respectively.

\noindent \textbf{Pruning error.} We investigate the pruning error accumulation issue. As depicted in Figure~\ref{fig: pruning_error}, our method exhibits slower error accumulation in comparison to Wanda, attributed to the block-wise pruning granularity.

\noindent \textbf{Impact of calibration dataset size.} We investigate the impact of the calibration set size on the accuracy of the pruned model without fine-tuning. Figure~\ref{fig: calibration_set_size} demonstrates the robustness of our method against variations in calibration set size. Even when the LLaMA-7B model is pruned with only one sample, the perplexities only increase by around 4.0 and 2.2 on the WikiText2 and PTB datasets, respectively. We provides more results in the supplementary materials.

\section{Conclusion}
In this study, we have proposed LLM-BIP, a block-wise structured pruning approach for large language models. LLM-BIP aims to prune attention heads in the MSA and channels in the FFN while minimizing the impact on the corresponding transformer block output. To this end, we have derived a simple pruning metric that efficiently and effectively measures the importance of connections block-wisely in a single forward pass. We have evaluated the efficacy of LLM-BIP on LLaMA models and the Vicuna-7B model using common zero-shot datasets. Our results have demonstrated that LLM-BIP achieves similar accuracy to existing structured pruning approaches at low sparsity levels while significantly outperforming them at high sparsity levels.

\paragraph{Limitation.} While the proposed LLM-BIP mitigates the error accumulation issue by employing a block-wise pruning scope, it remains a local pruning approach and may still encounter error accumulation.

\bibliography{CameraReady/main}

\end{document}